\documentclass[12pt]{article}

\usepackage{times}
\usepackage[numbers]{natbib}

\usepackage{amssymb,geometry,mathrsfs,mathtools,verbatim}
\usepackage{amsmath,bm,array,color,centernot}
\usepackage{indentfirst,float,enumerate}
\usepackage{tikz,lscape,multirow,graphicx,tabularx,booktabs}
\usepackage[noend]{algpseudocode}
\usepackage{algorithmicx}
\usepackage[Algorithm,ruled]{algorithm}
\usepackage{subcaption}

\usepackage[scr=euler,scrscaled=.96]{mathalfa}

\graphicspath{{example2/}}

\newenvironment{sciabstract}{%
\begin{quote} \it}
{\end{quote}}

\newcounter{lastnote}

\newcolumntype{Y}{>{\centering\arraybackslash}X}

\makeatletter
\def\BState{\State\hskip-\ALG@thistlm}
\makeatother

\makeatletter
\newcommand{\DESCRIPTION@original@item}{}
\let\DESCRIPTION@original@item\item
\newcommand*{\DESCRIPTION@envir}{DESCRIPTION}
\newlength{\DESCRIPTION@totalleftmargin}
\newlength{\DESCRIPTION@linewidth}
\newcommand{\DESCRIPTION@makelabel}[1]{\llap{#1}}%
\newcommand{\DESCRIPTION@item}[1][]{%
  \setlength{\@totalleftmargin}%
       {\DESCRIPTION@totalleftmargin+\widthof{\textbf{#1 }}-\leftmargin}%
  \setlength{\linewidth}
       {\DESCRIPTION@linewidth-\widthof{\textbf{#1 }}+\leftmargin}%
  \par\parshape \@ne \@totalleftmargin \linewidth
  \DESCRIPTION@original@item[\textbf{#1}]%
}

\makeatother

\newcounter{theorem}
\newtheorem{theorem}{Theorem}[section]
\newtheorem{remark}[theorem]{Remark}

\newtheorem{procedure}{Procedure}

\title{On the EM-Tau Algorithm: a new EM style algorithm with partial E-steps}

\author
{Val Andrei Fajardo,$^{1,2\ast}$ Jiaxi Liang$^{1}$\\
\\
\normalsize{$^{1}$Precima Inc.,}
\normalsize{438 University Avenue, Toronto, ON, Canada}\\
\normalsize{$^{2}$Integrate.ai Inc.,}
\normalsize{23 Sultan Street, Toronto, ON, Canada}\\
\normalsize{$^\ast$To whom correspondence should be addressed; E-mail:  andrei@integrate.ai}
}

\date{}

\begin{document}




\maketitle


\begin{sciabstract}
The EM algorithm is one of many important tools in the field of statistics.
While often used for imputing missing data, its widespread applications include
other common statistical tasks, such as clustering. In clustering, the EM
algorithm assumes a parametric distribution for the clusters, whose parameters
are estimated through a novel iterative procedure that is based on the theory of
maximum likelihood. However, one major drawback of the EM algorithm, that renders
it impractical especially when working with large datasets, is that it often
requires several passes of the data before convergence. In this paper, we introduce
a new EM-style algorithm that implements a novel policy for performing partial E-steps. We call the new algorithm the EM-Tau algorithm, which can approximate the traditional EM algorithm with high accuracy but with only a fraction of the running time.
\\

\normalfont{\noindent \textbf{Keywords:} EM Algorithm; clustering; unsupervised
learning; big data; Gaussian mixture models.}
\end{sciabstract}


\section{Introduction}

With the dramatic progress and development of data collection and storage technologies, we nowadays find ourselves surrounded by a massive volume of information and data from across many different sources (e.g., webpages, social media platforms, videos, internet of things, etc.). There is indeed large potential to gain valuable insights from big data; however, obtaining such insights through traditional statistical methods are often precluded by several computational challenges. One example is the expectation maximization (EM) algorithm, which is a classical tool in statistics that was first introduced by Dempster et al.\ in \cite{dempster1977maximum}, and is the central focus of this paper. The EM algorithm represents an interative procedure, where a single iteration is composed of an E-step and an M-step, that finds the maximum likelihood (ML) or maximum a posteriori (MAP) estimates of parameters of a specified statistical model. It has been widely applied to many statistical learning problems (of moderately sized datasets) with latent variables, such as the Gaussian mixture models (GMM) and hidden Markov models (e.g., see Amari \cite{amari1995information}, Bengio and Frasconi \cite{bengio1995input}, Xu and Jordan \cite{xu1993unsupervised}). A main reason for its widespread use is the fact that it is stable and has been mathematically proven to converge (e.g., see Wu \cite{wu1983convergence}, Boyles \cite{boyles1983convergence}, Xu and Jordan \cite{xu1996convergence}, and Neal et al.\ \cite{neal1998view}). Specifically, under some mild conditions, the EM algorithm is guaranteed to monotonically increase the observed data log likelihood at each iteration until it reaches a local maximum (or a saddle point). 

Despite its overall simplicity in implementation and its attractive convergence properties, the EM algorithm struggles when dealing with larger datasets -- a major criticism of the EM algorithm is that it often converges slowly, requiring several passes of the dataset. To combat this slow convergence problem, researchers in this area have developed various variations of the traditional EM algorithm, which accelerate convergence through certain modifications to either the E-step, M-step, or both. We conclude this introduction by briefly discussing some of these variations.

One way to accelerate the convergence of the EM algorithm is by modifying the M-step. A popular class of EM variants is called the ``generalized EM'' (GEM). The GEM performs the E-step in the usual way, but the M-step only partially. That is, instead of maximizing the conditional likelihood function in the M-step, it finds some values of the parameters that simply increase the current iteration's likelihood. A typical example in this class is the ECM algorithm (e.g., see Meng and Robin \cite{meng1993maximum}) in which the M-step maximizes each parameter individually via a sequence of conditional maximization steps. Another special example in the GEM class that is worth mentioning is the SAGE algorithm (Fessler and Hero \cite{fessler1994space}), which additionally makes use of the idea of data reduction in its partial M-step. Rather than maximizing all parameters by using the complete data space, with the SAGE algorithm, each parameter is updated sequentially by alternating between several small hidden-data spaces. Other GEM variants of note can be found in the papers by Meng and Rubin \cite{meng1992recent}, Liu and Rubin \cite{liu1994ecme}, Yin et al.\ \cite{yin2012accelerating}, and Meng and Van Dyk \cite{meng1997algorithm}. Besides the GEM, other EM variants that aim to accelerate the maximization of the conditional likelihood function in the M-step have been researched, such as: quasi-Newton's method (e.g., see Lange \cite{lange1995quasi}), conjugate gradient (e.g., see Jamshidian and Jennrich \cite{jamshidian1997acceleration}), parameter expansion (e.g., see Liu et al.\ \cite{liu1998parameter}), and sufficient statistics (e.g., see Bradley et al.\ \cite{bradley1998scaling}, Ordonez and Omiecinski \cite{ordonez2002frem}). 

Another means to speed up the EM algorithm is through the modification of the E-step, which is especially beneficial when the data size is massive. The key idea is to identify a set of ``good'' points whose membership probabilities do not need to be re-evaluated at each iteration. Therefore, the E-step is performed only partially with a subset of the complete data to reduce the computational cost. Two typical examples are the Sparse EM (e.g., see Neal and Hinton \cite{neal1998view}) and the Lazy EM (e.g., see Thiesson et al.\ \cite{thiesson2001accelerating}). These methods have been theoretically justified by Neal and Hinton in \cite{neal1998view}, where the authors proved that as long as the data points are visited regularly, the algorithm still maintains the convergence property of the traditional EM. A similar but more aggressive method in terms of identifying good points is the EM* algorithm, which was introduced by Kurban et al.\ in \cite{kurban2016algorithm}. There, the data space for the E-step is reduced monotonically at each iteration, leading to an approximation of the traditional EM result. Lastly, it is straightforward to see that the E-step (or the partial E-step) is perfectly suitable for implementing the distributed and parallel computing techniques. For work in this area, we refer the reader to the papers by Chu et al.\ \cite{chu2006map}, Wolfe et al.\ \cite{wolfe2008fully}, Kurasova et al.\ \cite{kurasova2014strategies}, and Kurban et al.\ \cite{kurban2016algorithm}.

In this paper we introduce a new policy for performing a partial E-step, which leads to a significant reduction in computational burden while still achieving comparable performance to that of the traditional EM algorithm. The rest of the paper is organized as follows. In section 2, we briefly review the traditional EM algorithm and its application to the GMM. The new EM style algorithm with partial E-steps is introduced in section 3, and some of its important mathematical properties are also discussed. Numerical examples and real data applications are provided in section 4. Lastly, we offer some concluding remarks in section 5.

\section{The EM Algorithm and GMMs}

Assume we have a set of observed data $\bm{X}=\left\{ \bm{x}_{1},\ldots, \bm{x}_{N} \right\}$, with a set of hidden data $\bm{Z}=\left\{ z_{1},\ldots, z_{N} \right\}$. The goal is to maximize the observed data log-likelihood function, which is given by
\begin{align}
\ell(\bm{\Theta})=\sum_{n=1}^{N}\ln\left\{ \sum_{z_{n}}p(\bm{x}_{n},\, z_{n}|\bm{\Theta})\right\}.
\end{align} 

This is not always an easy task due to the presence of the summation inside the logarithm. Moreover, maximizing the complete data log-likelihood $\ell_{c}(\bm{\Theta}) = \sum_{n}\ln \, p(\bm{x}_{n},\, z_{n}|\bm{\Theta})$ is also impossible since $z_{n}$ is unobserved. The basic idea of the EM algorithm is then to find the maximum of $\ell(\bm{\Theta})$ via a two-step process, which we describe next. At each iteration, we first perform an E-step (i.e., ``expectation'' step) in which we build the expected complete log-likelihood with respect to the conditional distribution of $z$ given observed data and the current estimate of the parameter. This conditional expectation is usually denoted as $Q(\bm{\Theta}|\bm{\Theta}^{[t]})$ and is given by
\begin{align}
\label{qfunction}
Q(\bm{\Theta}|\bm{\Theta}^{[t]}) = \rm{E}_{\bm{Z}|\bm{X},\bm{\Theta}^{[t]}}(\ell_{c}(\bm{\Theta})).
\end{align}
Next, we perform the so-called M-step (i.e., ``maximization'' step), during which we update the parameter estimates by finding the values that maximize the $Q$ function (i.e., Equation \eqref{qfunction}) obtained from the E-step. That is,
\begin{align}
\label{mstep}
\bm{\Theta}^{[t+1]} = \underset{\bm{\Theta}}{\arg\max} Q(\bm{\Theta}|\bm{\Theta}^{[t]}).
\end{align}
The E-step and M-step are performed successively and iteratively until the specified convergence criteria has been met (e.g., the observed log likelihood no longer increases).

The EM algorithm is a natural solution to fit a GMM, where the observed data are assumed to be generated from a mixture of several Gaussian distributions, and the true label of each data point is a hidden variable. In what follows, we briefly review the traditional EM algorithm for the GMM, and motivate the need to develop an approximation to traditional EM. 

The GMM is an intuitive and useful tool for clustering (e.g., see McLachlan et al.\ \cite{mclachlan1988mixture}). In the context of clustering, $K$ Gaussian components represent $K$ clusters, and the label $z_n$ represents the true membership of the point $\bm{x}_n$, that is, $\mathrm{Pr}(z_n=k)=\pi_k$. Without knowing the true values of the labels, we can write the joint log-likelihood function of observed data $\bm{X}=\left\{ \bm{x}_{1},\ldots, \bm{x}_{N} \right\}$ as
\begin{align}
\label{gmm}
\ell(\bm{\Theta}|\bm{X})=\sum_{n=1}^{N}\ln\left\{ \sum_{k=1}^{K} \pi_k \cdot \Phi(\bm{x}_{n}|\bm{\mu}_{k},\,\bm{\Sigma}_{k})\right\},
\end{align}
where $\Phi(\cdot|\bm{\mu}_{k},\,\bm{\Sigma}_{k})$ is the density of the $k$-th Gaussian component $\mathscr{N}(\bm{\mu}_{k},\,\bm{\Sigma}_{k})$, which we assume is of dimension $d$. The unknown parameters to be estimated are thus $\bm{\Theta}=\left\{ (\bm{\mu}_{k},\,\bm{\Sigma}_{k},\,\pi_k)\right\}_{k=1}^K $. For each observed point $\bm{x}_n$, we obtain the membership probability to the $k$-th component by a straightforward application of Bayes' theorem. Specifically, we have that
\begin{align}
w_{k}(\bm{x}_n, \bm{\Theta})=\mathrm{Pr}(z_n=k|\bm{x}_n,\, \bm{\Theta})=\frac{\pi_k \cdot \Phi(\bm{x}_{n}|\bm{\mu}_{k},\,\bm{\Sigma}_{k})}{\sum_{j=1}^{K}\pi_j \cdot \Phi(\bm{x}_{n}|\bm{\mu}_{j},\,\bm{\Sigma}_{j})}.
\end{align}
We then say that $\bm{x}_n$ is clustered into the component for which it has the highest membership probability. 

To fit the GMM with the EM algorithm, the $Q$ function can be written as follows:
\begin{align}
Q(\bm{\Theta}|\bm{\Theta}^{[i-1]}) &= {\rm{E}}_{\bm{Z}|\bm{X},\bm{\Theta}^{[i-1]}}(\ell_{c}(\bm{\Theta})) \nonumber \\
&= \sum_{n=1}^{N}{\rm{E}}\left\{ \ln \left[ \prod_{k=1}^{K}(\pi_{k}\cdot \Phi(\bm{x}_{n}|\bm{\mu}_{j},\,\bm{\Sigma}_{j}))^{I(z_{n}=k)} \right] \right\} \nonumber \\
&= \sum_{n=1}^{N}\sum_{k=1}^{K}{\rm{E}}\left[ I(z_{n}=k)|\bm{x}_{n}, \bm{\mu}^{[i-1]}_{j},\,\bm{\Sigma}^{[i-1]}_{j} \right] \cdot \ln \left[ \pi_{k}\cdot \Phi(\bm{x}_{n}|\bm{\mu}_{j},\,\bm{\Sigma}_{j}) \right] \nonumber \\
&= \sum_{n=1}^{N}\sum_{k=1}^{K} w_{k}(\bm{x}_n, \bm{\Theta}^{[i-1]}) \cdot \ln \left[ \pi_{k}\cdot \Phi(\bm{x}_{n}|\bm{\mu}_{j},\,\bm{\Sigma}_{j}) \right].
\end{align}
Therefore, the main purpose of the E-step (in the $i$-th iteration) is to evaluate the membership probabilities based on the current estimates of parameters by using
\begin{align}
w^{[i]}_{k}(\bm{x}_n)=w_{k}(\bm{x}_n, \bm{\Theta}^{[i-1]})=\frac{\pi^{[i-1]}_k \cdot \Phi(\bm{x}_{n}|\bm{\mu}^{[i-1]}_{k},\,\bm{\Sigma}^{[i-1]}_{k})}{\sum_{j=1}^{K}\pi^{[i-1]}_j \cdot \Phi(\bm{x}_{n}|\bm{\mu}^{[i-1]}_{j},\,\bm{\Sigma}^{[i-1]}_{j})}.
\end{align}
In the M-step, there exists explicit closed forms of the maximal parameter estimates. In particular, we update the Gaussian parameters as follows:
\begin{align}
& \bm{\mu}^{[i]}_{k}=\frac{\sum^{N}_{n=1}w^{[i]}_{k}(\bm{x}_n)\cdot \bm{x}_n}{\sum^{N}_{n=1}w^{[i]}_{k}(\bm{x}_n)}\\
& \bm{\Sigma}^{[i]}_{k}=\frac{\sum^{N}_{n=1}w^{[i]}_{k}(\bm{x}_n)\cdot (\bm{x}_{n}-\bm{\mu}^{[i]}_{k})(\bm{x}_{n}-\bm{\mu}^{[i]}_{k})}{\sum^{N}_{n=1}w^{[i]}_{k}(\bm{x}_n)}\\
& \pi^{[i]}_{k}=\frac{1}{N}\sum^{N}_{n=1}w^{[i]}_{k}(\bm{x}_n).
\end{align}

The E-step and the M-step are then repeatedly performed until the estimates of the Gaussian parameters converge to a local optima (or saddle point). The procedure of the traditional EM algorithm is summarized in Algorithm \ref{pcode-traditional}.
 
\begin{algorithm}[htbp]
\caption{Traditional EM Algorithm}\label{pcode-traditional}
\begin{algorithmic}[1]
\State \%\% Initialization
\State $\bm{\Theta}^{[0]} \gets \mathit{Initialization()}$
\State $i \gets 1$
\State \%\% Perform E-Step and M-Step until convergence
\While {!convergence}
\State $\bm{w}^{[i]} \gets \mathit{EStep}(\bm{\Theta}^{[i-1]})$
\State $\bm{\Theta}^{[i]} \gets \mathit{MStep(\bm{w}^{[i]})}$
\State $\text{convergence} \gets \mathit{isCloseEnough}(\bm{\Theta}^{[i]},\bm{\Theta}^{[i-1]})$
\State $i \gets i + 1$
\EndWhile
\end{algorithmic}
\end{algorithm}

It bears mentioning that in each iteration we need to evaluate $K$ Gaussian densities for $N$ points in the E-step, and that this scales as $O(KNd^3)$. Moreover, the M-step requires $O(KNd^2)$ work in order to update the estimates of the Gaussian parameters. Although the time complexity is linear in both sample size and number of clusters for each iteration, computational challenges nevertheless arise when the sample size is too massive or when several iterations are required before convergence.

\section{The  EM algorithm with partial E-steps}
As mentioned, the traditional EM algorithm is not a practical choice for clustering points from a massive dataset due to the potential for slow convergence. In this section, we describe a general framework for approximating the EM algorithm. The underlying intuition of this framework is that some data points are more bound to a
single cluster than others, and that these points are less likely to yield significant changes in their
membership values throughout the evolution of the algorithm. Hence, a sensible
idea would be to ignore the membership updates of such points; which in turn, would enable the algorithm
to focus its efforts on the data points that lie on the boundaries of multiple clusters.
We carefully describe the framework next.

First of all, at each iteration, the dataset is split into two types of points, namely: (i) \emph{active} points
whose membership weights are to be updated in the current iteration; and (ii) \emph{non-active}
points whose membership weights are not to be updated. After the active points have been identified, the E-step is performed on only these points (i.e., the membership
weights of non-active points remain the same as they were in the previous iteration). Finally,
to complete the current iteration of the algorithm, the M-step is performed on the
entire membership matrix. Variants of the EM algorithm, which implement this framework,
have come to be known as EM algorithms with a partial E-step.

By letting $\mathcal{A}^{[i]}_X$ represent the set of active points for the $i$-th iteration, we can write the pseudocode for an EM algorithm with partial E-steps as given by Algorithm \ref{pcode-general}, where $\sigma^{[i]}$ represent the historical information of the EM algorithm up
to (and including) the $i$-th iteration and $f(\cdot)$ is a \emph{procedure} that determines the set of
active points.

\begin{algorithm}[htbp]
\caption{EM Algorithm with a partial E-step}\label{pcode-general}
\begin{algorithmic}[1]
\State \%\% Initialization
\State $\bm{\Theta}^{[0]} \gets \mathit{Initialization()}$
\State $i \gets 1$
\State \%\% Perform E-Step (on active points) and M-Step until convergence
\While {!convergence}
\State $\bm{w}^{[i]}[\mathcal{A}_X^{[i]}] \gets \mathit{EStep}(\bm{\Theta}^{[i-1]},\mathcal{A}_X^{[i]})$
\%\% update membership of active points
\State $\bm{w}^{[i]}[\bm{X}\setminus \mathcal{A}_X^{[i]}] \gets \bm{w}^{[i-1]}[\bm{X}\setminus \mathcal{A}_X^{[i]}]$
\State $\bm{\Theta}^{[i]} \gets \mathit{MStep(\bm{w}^{[i]})}$
\State $\text{convergence} \gets \mathit{isCloseEnough}(\bm{\Theta}^{[i]},\bm{\Theta}^{[i-1]})$
\State $\mathcal{A}^{[i+1]}_X \gets f(\sigma^{[i]})$ \%\% identify active points for next iteration
\State $i \gets i + 1$
\EndWhile
\end{algorithmic}
\end{algorithm}

\subsection{The EM-Tau algorithm}

We next introduce a new EM-style algorithm with a partial E-step, which we refer to as the EM-Tau algorithm. The fashion in which the active points are determined by this new algorithm is described in Procedure 1.

\begin{procedure}[$f_\tau(\cdot)$]
For each point $x \in \bm{X}$, let $c_j$ denote the cluster to which $x$ belongs after the $j$-th E-step.
Moreover, let $m_j$ represent a counter variable, which counts the number of
previous iterations where $x$ has consecutively belonged to $c_j$. That is, for $j>0$,
\begin{equation}
m_j = \left\{
\begin{array}{l l}
m_{j-1} + 1, & c_j = c_{j-1}\\
1, & c_j \neq c_{j-1}
\end{array}
\right..
\end{equation}
Note that we define $m_0=0$. A simple procedure for determining $\mathcal{A}^{[i+1]}_X$
can then be described as follows:
\begin{enumerate}

\item After the $i$-th E-step is performed, determine $(c_i,m_i$) for each point
in $x\in\mathcal{A}_X^{[i]}$;

\item Then, the set of active points for the next iteration is given by
$\mathcal{A}_X^{[i+1]} = \{x\in A_X^{[i]}: m_i < \tau \}$, where
$\tau$ is an integer-valued parameter such that $\tau > 0$.

\end{enumerate}
\end{procedure}
Simply put, with the EM-Tau algorithm, a point becomes non-active once it has remained in the same
cluster for $\tau$ consecutive iterations. 

\begin{remark} 
The EM-Tau algorithm can be viewed as an approximation to the traditional EM algorithm, where the parameter $\tau$ controls the accuracy of the approximation. In particular, by setting $\tau=\infty$,
the EM-Tau algorithm becomes exactly equivalent to the traditional EM algorithm.
\end{remark}

\begin{remark}
The EM-Tau algorithm may terminate after the $t$-th iteration in either one of the following two ways: (i) if the estimates of the parameters at time $t$ satisfy the convergence criteria, or (ii) if the set of active points $\mathcal{A}_X^{[t+1]}$ is an empty set. 
\end{remark}

\subsubsection{Convergence of the EM-Tau Algorithm}

The purpose of the current section is to provide a theoretical justification of the EM-Tau algorithm. In what follows, we prove that if the set of active points is non-empty at the completion of the EM-Tau algorithm, then it must be that the parameter estimates has converged to a local stationary point of the observed log-likelihood function.

First of all, note that a lower bound for the observed log-likelihood function can be obtained by Jensen's inequality. Specifically,
\begin{align}
\label{bound}
\ell(\bm{\Theta}) &= \sum^{N}_{n=1} \left\{ \ln \left[ \sum_{z_{n}}p(\bm{x}_{n},\,z_{n}|\bm{\Theta}) \right] \right\} \nonumber \\
&= \sum^{N}_{n=1} \left\{ \ln \left[ \sum_{z_{n}}q(z_{n})\frac{p(\bm{x}_{n},\,z_{n}|\bm{\Theta})}{q(z_{n})} \right] \right\}\nonumber \\
& \geq \sum^{N}_{n=1} \sum_{z_{n}} \left\{q(z_{n})\cdot \ln \left[ \frac{p(\bm{x}_{n},\,z_{n}|\bm{\Theta})}{q(z_{n})} \right] \right\} \nonumber \\
&= \sum^{N}_{n=1}\left\{ {\rm{E}}_{q} \left[ \ln p(\bm{x}_{n},\,z_{n}|\bm{\Theta}) \right]  + {\rm{H}}(q) \right\},
\end{align}
where $q(z)$ is an arbitrary distribution of the hidden variable $z$, and ${\rm{H}}(q)$ is the entropy of distribution $q$. For simplicity, we define 
\begin{align*}
F_{n}(\bm{\Theta},\, q) = {\rm{E}}_{q}\left[ \ln p(\bm{x}_{n},\,z_{n}|\bm{\Theta}) \right] + {\rm{H}}(q), 
\end{align*}
and 
\begin{align*}
F(\bm{\Theta},\, q)=\sum_{n}F_{n}(\bm{\Theta},\, q)
\end{align*}
so that Equation \eqref{bound} can be re-expressed as $\ell(\bm{\Theta}) = F(\bm{\Theta}, q)$. Neal and Hinton in \cite{neal1998view} show that the traditional EM algorithm can be equivalently expressed as a maximization-maximization procedure in terms of $F(\bm{\Theta},\, q)$. That is, the purpose of the E-step is to find the optimal distribution $q$ based on the current estimate of parameters $\bm{\Theta}$, while the purpose of the M-step is to find the optimal parameter $\bm{\Theta}$ based on the current distribution $q$. Specifically, we write:
\begin{align*}
& {\text{E-step: }}\,  q^{[t+1]}=\underset{q}{\arg\!\max}\,F(\bm{\Theta}^{[t]},\, q)\\
& {\text{M-step: }}\, \bm{\Theta}^{[t+1]}=\underset{\bm{\Theta}}{\arg \!\max}\,F(\bm{\Theta},\, q^{[t+1]}).
\end{align*}

Furthermore, Neal and Hinton in \cite{neal1998view} also prove that if a local maximum of $F$ occurs at $(\bm{\Theta}^{\star},q^{\star})$, then the log-likelihood $\ell(\bm{\Theta})$ must also achieve a local maximum at $\ell(\bm{\Theta}^{\star})$ (e.g., see Theorem 2 in Neal and Hinton \cite{neal1998view}). Ultimately, as long as the $F$ function is improved at each of the individual E-steps and M-steps, the procedure is guaranteed to find a local maximum of the observed data log likelihood.

In the EM-Tau procedure, the M-step is the same as the traditional EM so that the $F$ function is obviously improved in the M-step. On the other hand, there is slightly more work needed to be done in order to verify the improvement of $F$ in the new partial E-step. First, let the conditional distribution of $z_{n}$ at the $[t+1]$-th iteration be $q_{n}^{[t+1]}(z_{n})$. It is important to realize that in the EM-Tau algorithm, we only update the conditional distribution for active points in the E-step. That is,
\begin{align*}
q_{n}^{[t+1]}(z_{n}) = 
\begin{cases} p(z_{n}|\bm{x}_{n},\,\bm{\Theta}^{[t]}) & \bm{x}_{n} \in \mathcal{A}^{[t]}_X \\ 
q_{n}^{[t]}(z_{n}) & \bm{x}_{n} \notin \mathcal{A}^{[t]}_X
\end{cases}.
\end{align*}
Or equivalently,  
\begin{align*}
q_{n}^{[t+1]}(z_{n}) = 
\begin{cases} \underset{q}{\arg\!\max}\,F_{n}(\bm{\Theta}^{[t]},\, q) & \bm{x}_{n} \in \mathcal{A}^{[t]}_X \\ 
q_{n}^{[t]}(z_{n}) & \bm{x}_{n} \notin \mathcal{A}^{[t]}_X
\end{cases},
\end{align*}
from which we obtain that
\begin{align*}
F(\bm{\Theta}^{[t]},\, q^{[t+1]}) - F(\bm{\Theta}^{[t]},\, q^{[t]}) = \sum_{\bm{x}_n\in\mathcal{A}^{[t]}_X}\left\{ F_{n}(\bm{\Theta}^{[t]},\, q^{[t+1]})- F_{n}(\bm{\Theta}^{[t]},\, q^{[t]}) \right\} \geq 0.
\end{align*}

Thus, the partial E-step in the EM-Tau algorithm is also guaranteed to improve the $F$ function so long as the active points set ${A}^{[t]}_X$ is non-empty. It then follows, from Theorem 2 of Neal and Hinton \cite{neal1998view}, that if the estimates of parameters converge before the set of active points empties, then the EM-Tau algorithm has found a local stationary point of the observed log-likelihood function.

\subsection{Other EM algorithms with partial E-steps}

In this section, we review two other previously developed EM algorithms that use a partial E-step, namely: the EM* and EM-Lazy algorithms.

\subsubsection{EM*}
The EM* algorithm was developed by Kurban et al.\ in \cite{kurban2016algorithm}. Its procedure for identifying active points is given in Procedure 2.

\begin{procedure}[$f_*(\cdot)$]
To determine the set $\mathcal{A}^{[i+1]}_X$, the EM* algorithm implements a
procedure, denoted by $f_*()$, which performs the following steps:
\begin{enumerate}
\item Determine the clusters to which each $x_n \in \mathcal{A}_X^{[i]}$ belongs;
\item For each cluster $k \in K$, generate a heap using the cluster-$k$ membership weights
of only those data points that belong to it and that are active;
\item $\mathcal{A}^{[i+1]}_X$ is then the set of points lying in the
tails associated with each of the $K$ heaps created in step 2.
\end{enumerate}
\end{procedure}

The intuition behind this procedure for identifying active points is that
points that are more bound to a cluster should have relatively larger membership weights for
it. Thus, generating a heap data structure for each cluster provides a fast--in linear
time--organization of the points belonging to them (i.e., points with larger membership
weights are near the head of the heap, while points with lesser membership values are found
near its tail.) For more details of this procedure, we refer the reader to the original
paper by Kurban et al.\ \cite{kurban2016algorithm}.

\subsubsection{EM-Lazy}

The EM-Lazy algorithm was first introduced by Thiesson et al.\ in \cite{thiesson2001accelerating}. We describe the procedure for determining active points in the EM-Lazy algorithm in Procedure 3.

\begin{procedure}[$f_{L}(\cdot)$]
To determine the set $\mathcal{A}^{[i+1]}_X$, the EM-Lazy algorithm implements a
procedure, denoted by $f_{L}()$, which performs the following steps:
\begin{enumerate}
\item Based on the pre-specified schedule, determine whether the current iteration performs a full E-step or a lazy E-step. If it's a full E-step then $\mathcal{A}^{[i+1]}_X = \bm{X}$ (i.e., the entire dataset), otherwise proceed to the next step;
\item For each point $x_{n}$, calculate $P_{n} = \underset{k}{\max} \left\{ w^{[i]}_{k}(x_{n}) \right\}$, which is the maximum membership probability;
\item Then $\mathcal{A}^{[i+1]}_X = \left\{ n|P_{n}>T \right\}$, where $T$ is a pre-specified threshold.
\end{enumerate}
\end{procedure}

The basic idea of the EM-Lazy procedure is similar to the one for EM-Tau, in that not all points are of equal importance at each iteration, and only a subset of points need to be updated in every E-step. The main difference lies in the criterion of identifying ``important'' points. The maximum membership probability calculated in step 2. of $f_L(\cdot)$ represents how strong a data point is associated with the assigned component. If $P_{n}$ is larger than a predetermined threshold, then the point $x_{n}$ is believed to be close enough to the center of the assigned component, and does not need any update in the next iteration. Another key difference is the additional revisiting mechanism in the EM-Lazy algorithm, where a predetermined schedule specifies when an iteration performs a full E-step (i.e., usually after several iterations with lazy or partial E-steps). By sacrificing a little computational efficiency, the revisiting mechanism prevents points from being ``trapped'' in the wrong component. The performance of EM-Lazy depends very much on the choice of the threshold $T$, and the predetermined schedule for full E-steps. Choosing an appropriate value for $T$ is not a straightforward task, especially when the number of components $K$ is large.  

\subsection{Complexity analysis}
We begin with a simple time complexity analysis of the traditional EM algorithm.
Recall that during the E-step, the algorithm computes the membership weights of
all the data points for each of the $K$ clusters. Assuming that the best possible
run time to evaluate a single density is of order $O(d)$, it then follows that
the best-case run time for the E-step is $O(KNd)$. Applying similar arguments to
the M-step yields a best-case run time of $O(Kd)$ for this step; and thus, an overall
running time of $O(KNd + Kd) = O(KNd)$ at each iteration.

It is also clear from Algorithms \ref{pcode-traditional} and \ref{pcode-general} that
we must know the time complexity of the procedure $f(\cdot)$
in order to obtain the time complexity of the partial E-step framework.
Moreover, it is obvious that to guarantee the same time complexity as the traditional
algorithm, the procedure $f(\cdot)$ should be implemented in linear time $O(n)$.
In this case, the general framework will have a run time of $O(KNd + Kd + N) = O(KNd)$
at each iteration. We note here that both $f_*(\cdot)$ and $f_\tau(\cdot)$ are
procedures that have $O(n)$ time complexity.

\begin{remark}
Although the time complexities of the partial E-step framework and the traditional algorithm are the same (i.e., assuming
a linear-time procedure $f(\cdot)$), there is yet a significant reduction in actual running time
with the former, which is a direct result of the fact
that $\mathcal{A}^{[i+1]}_X \subseteq X $.
\end{remark}

In some instances, the partial E-step framework may also require more memory than the
traditional algorithm. For example, with the $f_\tau(\cdot)$ procedure,
the algorithm needs to carry the pair $(c_i, m_i)$ for each active point
$\bm{x}_n$ onto the next iteration. In most cases, one additional byte for $c_i$ and another
for $m_i$ should suffice (i.e., this would allow both $K$ and $\tau$ to be as great
as 255). As a result, the $f_\tau(\cdot)$ procedure requires $2n$ additional
bytes of memory in comparison to the tradtional EM algorithm. 

\section{Examples and Applications}
\subsection{Example 1: Two-Component GMM}
\begin{figure}[H]
\centering
\begin{subfigure}[t]{0.5\textwidth}
\includegraphics[scale=0.75]{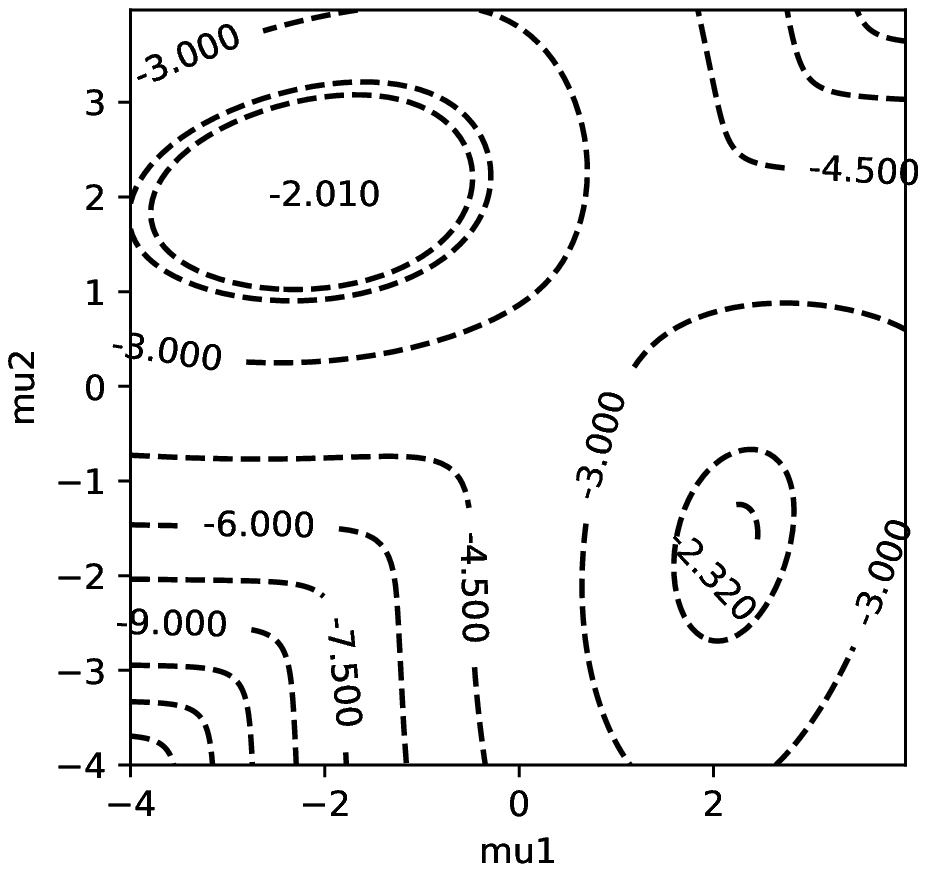}
\caption{True mean log-likelihood}
\label{true_contour}
\end{subfigure}%
\begin{subfigure}[t]{0.5\textwidth}
\includegraphics[scale=0.75]{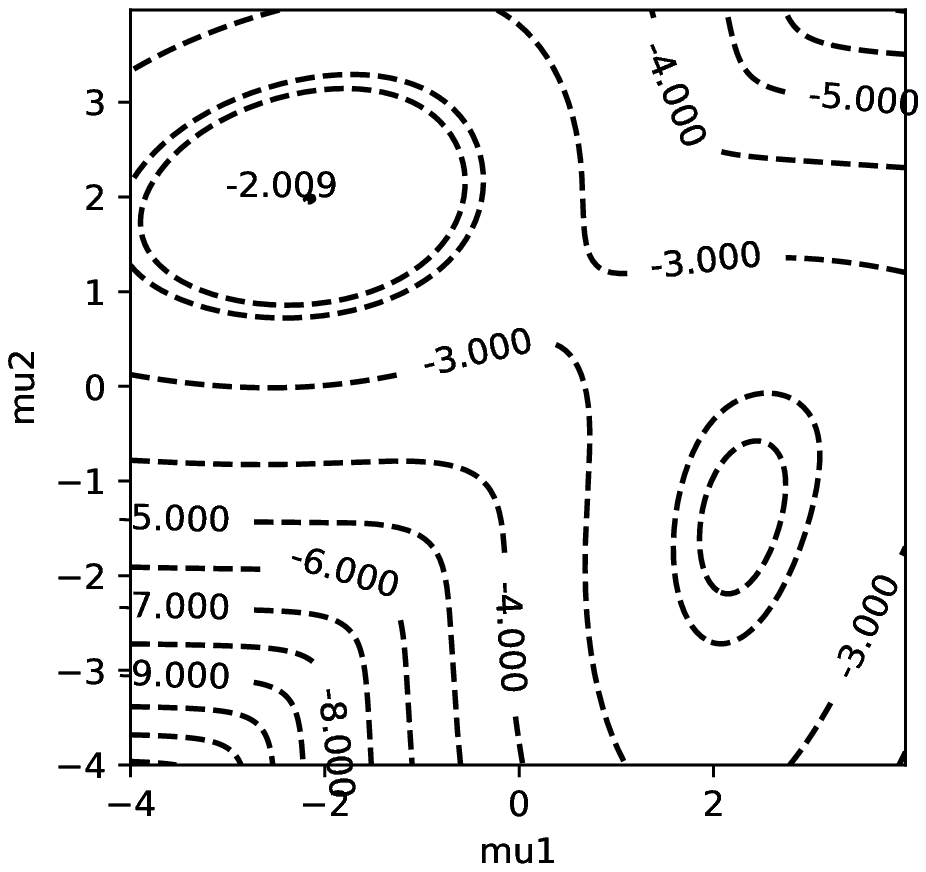}
\caption{Traditional EM}
\label{em_trad_contour}
\end{subfigure}
\end{figure}

The current example, which takes inspiration from the example found in the paper by Ueda and Nakano \cite{ueda1998deterministic}, serves as a simple illustration of the approximated log-likelihood functions when using various values of $\tau$ with the EM-Tau algorithm. In particular, we consider a two-component GMM whose density function is given by
\begin{equation}
p(x) = 0.3\Phi(x,-2,1) + 0.7\Phi(x,2,1),
\end{equation}
and from which we generate a random sample of $N=1000$ points. Figure \ref{true_contour} illustrates the contour plot of the true log-likelihood surface (i.e., specifically, $\ell(\bm{\Theta})/N$). 

\begin{figure}[H]
\centering
\begin{subfigure}[t]{0.33\textwidth}
\includegraphics[width=0.98\linewidth]{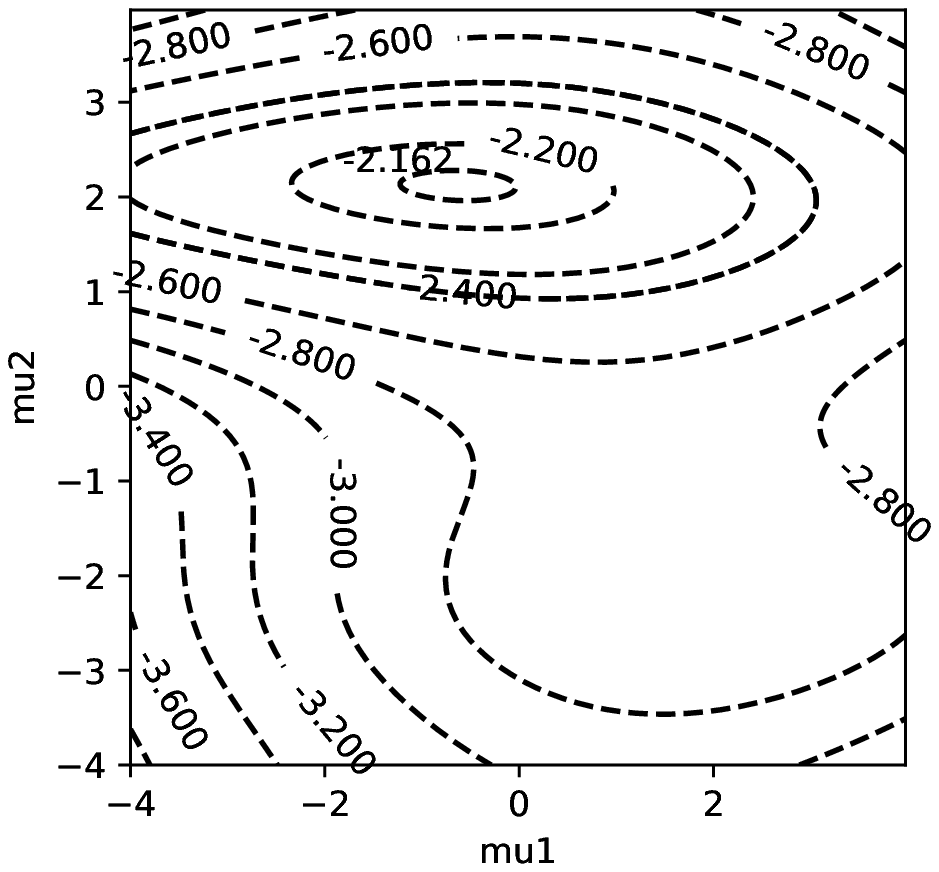}
\caption{$\tau=10$}
\end{subfigure}%
\begin{subfigure}[t]{0.33\textwidth}
\includegraphics[width=0.98\linewidth]{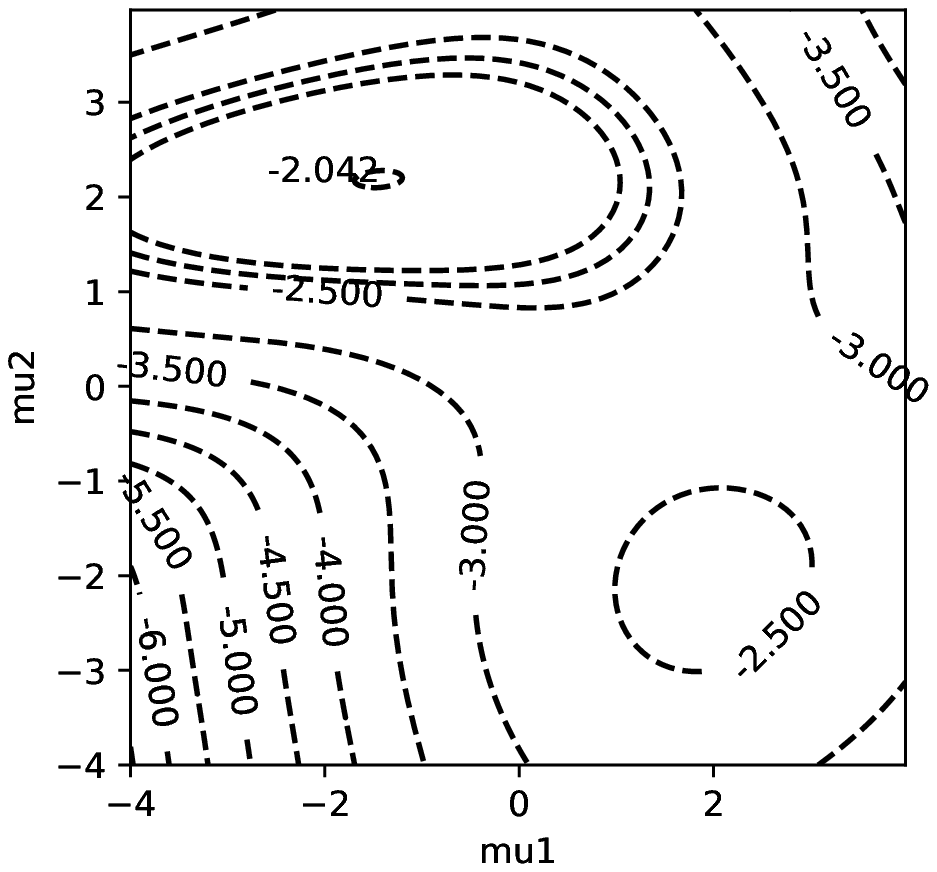}
\caption{$\tau=20$}
\end{subfigure}
\begin{subfigure}[t]{0.33\textwidth}
\includegraphics[width=0.98\linewidth]{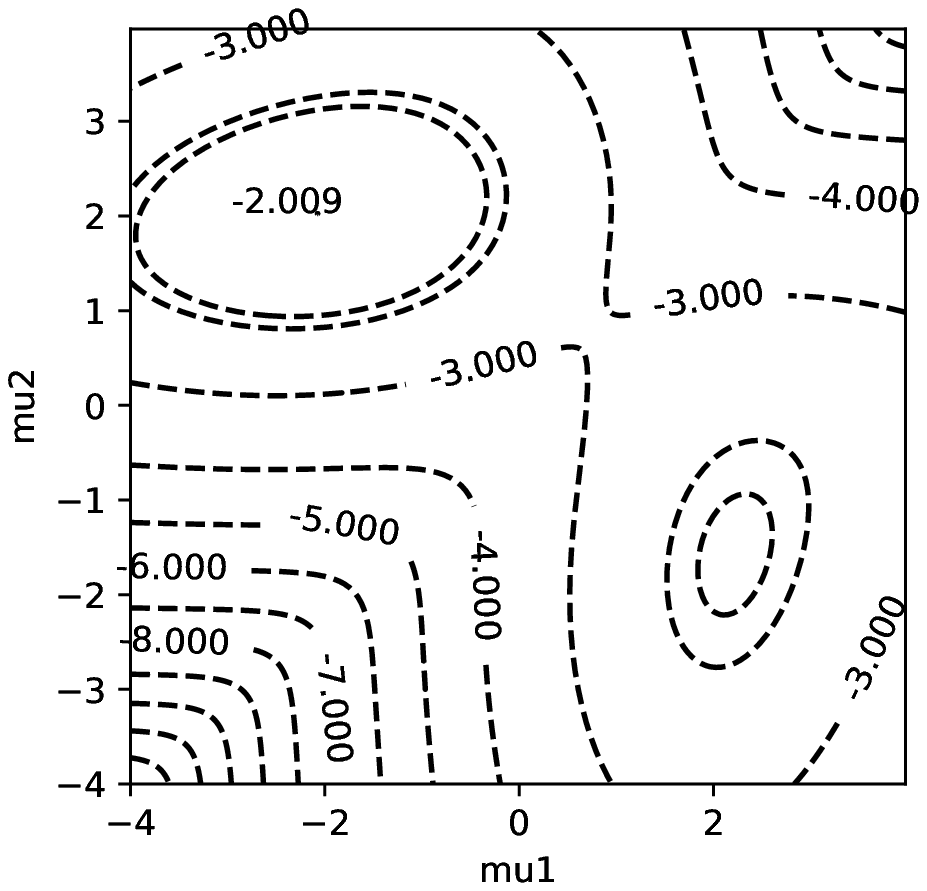}
\caption{$\tau=30$}
\end{subfigure}
\begin{subfigure}[t]{0.33\textwidth}
\includegraphics[width=0.98\linewidth]{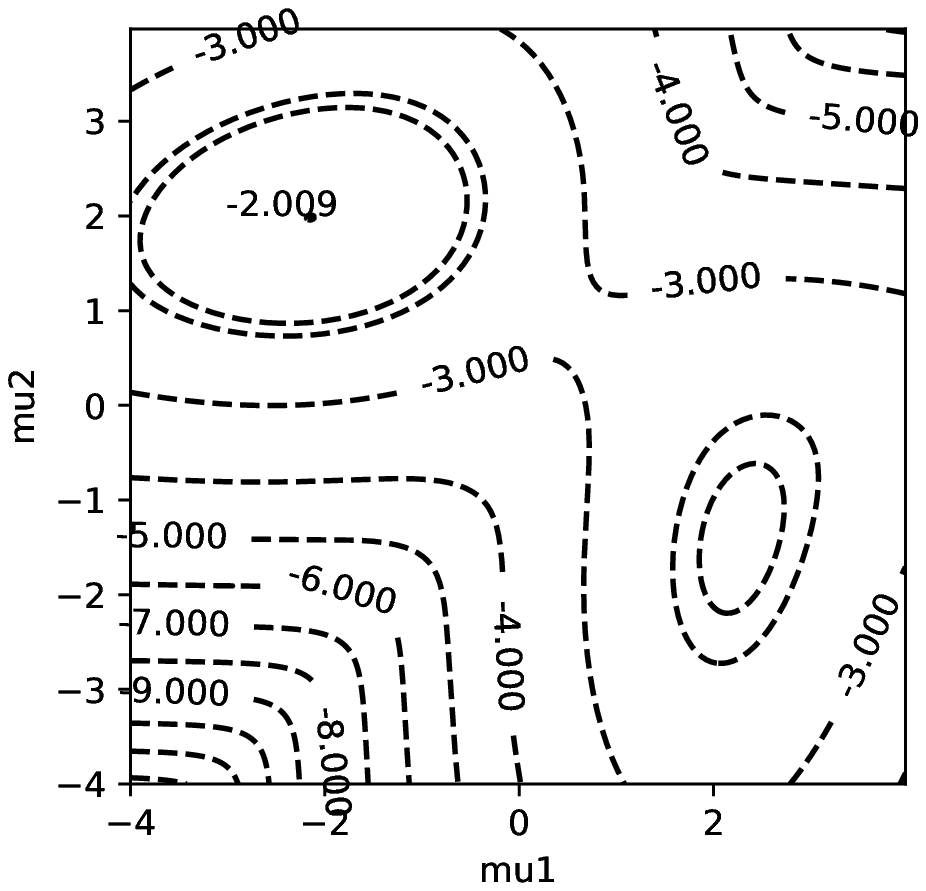}
\caption{$\tau=40$}
\end{subfigure}%
\begin{subfigure}[t]{0.33\textwidth}
\includegraphics[width=0.98\linewidth]{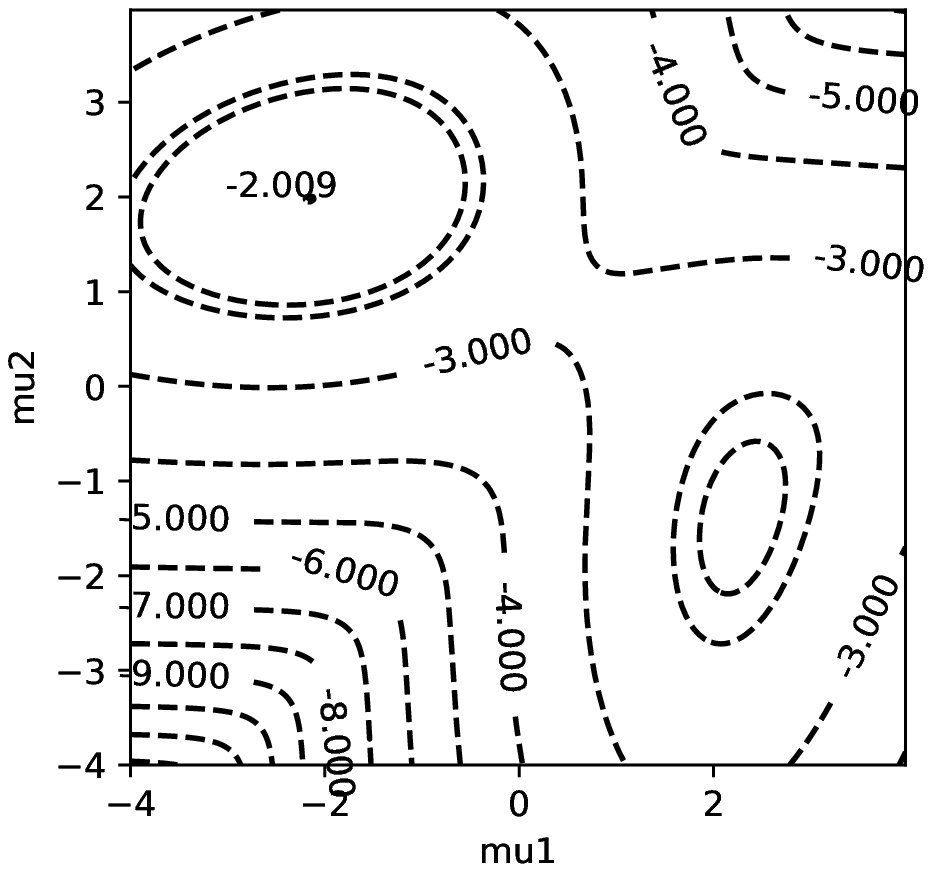}
\caption{$\tau=50$}
\end{subfigure}
\begin{subfigure}[t]{0.33\textwidth}
\includegraphics[width=0.98\linewidth]{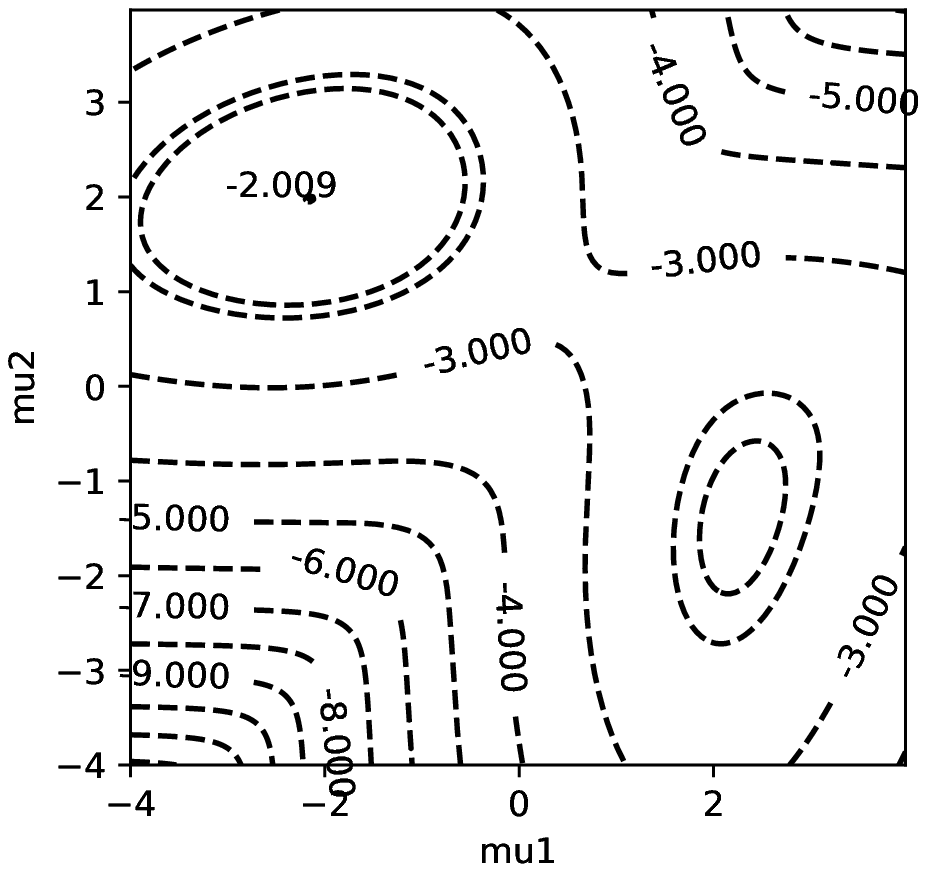}
\caption{$\tau=60$}
\end{subfigure}
\caption{Contour plots of log-likelihood surface resulting from EM-Tau}
\label{contour_emtau}
\end{figure}

For the sake of comparison, we performed the traditional EM algorithm on the 1000 sample observations, which converged after 64 iterations. The contour plot of the log-likelihood surface derived from the final parameter estimates of the traditional EM algorithm is depicted in Figure \ref{em_trad_contour}. As is evidenced by this plot, the traditional EM algorithm does quite well in this example. Contour plots of the log-likelihood surfaces obtained from the final parameter estimates of the EM-Tau algorithm with the various values of $\tau$ are provided in Figure \ref{contour_emtau}.

The final parameter values for each of the EM-Tau algorithms are reported in Table \ref{ex1-table}. We note that the EM-Tau algorithm with $\tau=30$ achieves nearly identical final parameter estimates as the traditional EM algorithm, but does so in slightly more than half the time.

\begin{table}[htbp]
\centering
\begin{tabular}{lrrr}
\toprule
         method &  iterations &  $(\alpha_1,\alpha_2,\mu_1,\mu_2,\sigma_1^2,\sigma_2^2)$ &  time(secs)  \\
\midrule
     EM-Tau(10) &          34 &          (0.46, 0.54, -0.92, 2.27,3.58, 0.8,) &   0.0145  \\
     EM-Tau(20) &          32 &          (0.36, 0.64, -1.69, 2.15, 1.82, 0.91) &   0.0259 \\
     EM-Tau(30) &          48 &          (0.30, 0.70, -2.06, 2.03, 1.08, 1.07) &  0.0352  \\
     EM-Tau(40) &          58 &          (0.30, 0.70, -2.12, 2.0, 0.99, 1.12) &  0.0448 \\
     EM-Tau(50) &          67 &          (0.29, 0.71, -2.12, 2.0, 0.98, 1.13) &  0.0510 \\
    EM-Tau(60) &          74 &          (0.29, 0.71, -2.12, 2.0, 0.98, 1.13) &  0.0589 \\
 EM-Traditional &          64 &               (0.29, 0.71, -2.12, 2.0, 0.98, 1.13)    &  0.0518  \\
\bottomrule
\end{tabular}
\caption{Final parameter estimates of EM-Tau and traditional EM algorithms for Example 1}
\label{ex1-table}
\end{table}

\subsection{Example 2: Handwritten Digits (12456)}

Our second example considers the classical MNIST Handwritten Digit dataset. While this dataset is often used in checking the performances of classification algorithms (i.e., the data is labelled), we emphasize that the truth labels were not used during the fitting of the GMM. Moreover, we only consider the images corresponding to the digits 1,2,4,5, and 6. This is because we are simply trying to compare the performances between the traditional EM algorithm and the new EM-Tau algorithm, and believe these digits to be the most distinguishable under an unsupervised learning setting.

We report the clustering results for each of the EM algorithms in Table \ref{MNIST-results}. There, the entries under the column ``membership error'' represent the Frobenius norm between the final membership weight matrices of the partial EM algorithm and that which results from the traditional EM algorithm. Furthermore, the entries under ``classification error'' represent the misclassification error when predicting the label of an image by the most frequently occurring label in its final cluster. As evidenced by the results in Table \ref{MNIST-results}, the parameter $\tau$ controls the trade-off between the computational efficiency and the accuracy of the approximation to the traditional EM algorithm. With an appropriately selected $\tau$, the EM-Tau significantly reduces the computational cost with almost no sacrifice on the accuracy. In this particular example, choosing $\tau=20$ results in 60\% speed up and only less than 0.2\% lost in the classification error.

\begin{table}[htbp]
\centering
\begin{tabular}{lrrrr}
\toprule
         method &  iterations &  membership error &  time(secs) &  classification error \\
\midrule
 EM-Traditional &          71 &                   &  287.200717 &       0.057394 \\
            EM* &           9 &          0.740303 &    8.521852 &       0.721261 \\
     EM-Tau(10) &          34 &          0.108597 &   66.946694 &       0.116596 \\
     EM-Tau(15) &          43 &          0.024865 &   98.208820 &       0.066564 \\
     EM-Tau(20) &          54 &          0.004719 &  117.375736 &       0.057327 \\
     EM-Tau(25) &          56 &          0.000602 &  134.949493 &       0.057227 \\
     EM-Tau(30) &          59 &          0.000100 &  152.897288 &       0.057294 \\
    EM-Tau(100) &          71 &          0.000000 &  288.784849 &       0.057394 \\
         Kmeans &             &                   &    2.680000 &       0.147184 \\
\bottomrule
\end{tabular}
\caption{Performance of the various cluster algorithms in Example 2}
\label{MNIST-results}
\end{table}

For comparisons sake, we also include the confusion matrices for the traditional EM, EM-Tau(25) and EM* algorithms in Tables 
\ref{conf-EM-trad}, \ref{conf-EM-tau-25}, and \ref{conf-EM-star}, respectively. The confusion matrix for Kmeans clustering is also reported in Table \ref{conf-Kmeans}. Unfortunately, the EM* algorithm failed to produce sensible approximations to the traditional EM algorithm during our tests. We believe that this likely due to the aggressive nature in which points becomes non-active. One idea which may
help to improve the performance of the EM* algorithm is to carry more points from each
heap on to the next iteration. For example, active points can be the collection of points in the lower
$x$-percentage of all the heaps, rather than just the tails. We note here that it was shown
by the authors in Kurban et al.\ \cite{kurban2016algorithm} that the median membership value often resided in (or near)
the tails of the heaps (e.g., one suggestion would be to use an $x > 50$). Another
possible improvement for the EM* algorithm would be to reduce the amount of active
points periodically rather than at every iteration (e.g., implement $f_*()$ every $m$-th iteration,
$m\geq 1$).

\begin{table}[htbp]
\centering
\begin{tabular}{lrrrrr}
\toprule
number &     1 &     2 &     4 &     5 &     6 \\
classification &       &       &       &       &       \\
\midrule
1              &  5669 &     0 &     1 &     0 &     0 \\
2              &   533 &  5931 &   148 &    47 &    86 \\
4              &   535 &    15 &  5675 &     2 &     7 \\
5              &     2 &     5 &     7 &  5363 &   297 \\
6              &     3 &     7 &    11 &     9 &  5528 \\
\bottomrule
\end{tabular}
\caption{Confusion matrix for EM-Traditional}
\label{conf-EM-trad}
\end{table}

\begin{table}[htbp]
\centering
\begin{tabular}{lrrrrr}
\toprule
number &     1 &     2 &     4 &     5 &     6 \\
classification &       &       &       &       &       \\
\midrule
1              &  5677 &     0 &     1 &     0 &     0 \\
2              &   524 &  5930 &   149 &    48 &    90 \\
4              &   536 &    16 &  5675 &     2 &     7 \\
5              &     2 &     5 &     6 &  5362 &   294 \\
6              &     3 &     7 &    11 &     9 &  5527 \\
\bottomrule
\end{tabular}
\caption{Confusion matrix for EM-Tau(25)}
\label{conf-EM-tau-25}
\end{table}

\begin{table}[htbp]
\centering
\begin{tabular}{lrrrrr}
\toprule
number &     1 &     2 &     4 &     5 &     6 \\
classification &       &       &       &       &       \\
\midrule
1              &  4911 &  2559 &  2926 &  2050 &  3301 \\
2              &  1814 &  3397 &  2895 &  3361 &  2611 \\
4              &    17 &     2 &    21 &    10 &     6 \\
\bottomrule
\end{tabular}
\caption{Confusion matrix for EM*}
\label{conf-EM-star}
\end{table}

\begin{table}[htbp]
\centering
\begin{tabular}{lrrrrr}
\toprule
number &     1 &     2 &     4 &     5 &     6 \\
classification &       &       &       &       &       \\
\midrule
1              &  6635 &   726 &   275 &  1017 &   466 \\
2              &    46 &  4626 &    22 &    29 &   136 \\
4              &     7 &   285 &  5425 &   386 &   215 \\
5              &    42 &   122 &    15 &  3859 &   163 \\
6              &    12 &   199 &   105 &   130 &  4938 \\
\bottomrule
\end{tabular}
\caption{Confusion matrix for sklearn.cluster.KMeans}
\label{conf-Kmeans}
\end{table}

\section{Discussion and future work}

In this paper, we introduced the EM-Tau algorithm for approximating the traditional EM algorithm through the partial E-step framework. Our numerical experiments illustrated that with an appropriate choice of parameter $\tau$, the EM-Tau algorithm can significantly improve the computational efficiency of traditional EM, with only negligible sacrifice in the accuracy.

On the subject of related future work, we believe that it is possible to combine the idea of the EM-Tau procedure with some other previously researched ideas for accelerating the EM algorithm. Of particular interest to the authors is the consideration of a revisiting mechanism with the EM-Tau algorithm. This mechanism can act as a safeguard that prevents the occurrence of the undesirable scenario where the set of active points becomes empty before the parameters converge. Another obvious direction would be to perform both a partial E-step and a partial M-step in each iteration. In both of the above mentioned potential modifications, the main challenge would be in finding a proper value for the tuning parameter $\tau$ such that a good balance between speed and accuracy is achieved. Moreover, we also believe that the benefits of the EM-Tau algorithm could be further realized if it were combined with the idea of deterministic annealing (e.g., see Ueda and Nakano in \cite{ueda1998deterministic}) in order to accelerate the convergence to a global optima of the log-likelihood surface. 

\bibliographystyle{acm}
\bibliography{references}
\end{document}